\documentclass[10pt]{article}

\usepackage{amsmath}
\usepackage{amssymb}

\usepackage{graphicx}

\usepackage{cite}

\usepackage{color} 

\usepackage{setspace} 
\usepackage[labelfont=bf,labelsep=period,justification=raggedright]{caption}

\makeatletter
\renewcommand{\@biblabel}[1]{\quad#1.}
\makeatother

\date{}

\begin{document}

\begin{flushleft}
{\Large
\textbf{A Generative Model of Natural Texture Surrogates}
}
\\
Niklas L\"udtke$^{1,2}$, 
Debapriya Das$^{1}$, 
 Lucas Theis$^{1}$, 
Matthias Bethge$^{1,2,3,\ast}$
\\
\bf{1} Werner Reichardt Center for Integrative Neuroscience and Institute of Theoretical Physics, University of T\"ubingen, T\"ubingen, Germany
\\
\bf{2} Bernstein Center for Computational Neuroscience T\"ubingen, T\"ubingen, Germany
\\
\bf{3} Max Planck Institute for Biological Cybernetics, T\"ubingen, Germany
\\
$\ast$ E-mail: matthias@bethgelab.org
\end{flushleft}

\section*{Abstract}

Natural images can be seen as patchworks of different textures for which local image statistics are roughly stationary within confined regions but otherwise can be highly diverse. In order to model the natural variety of textures, we sampled $64 \times 64$ patches of homogeneous textures from a large image database and described each patch by a set of texture parameters obtained with a popular texture algorithm. Texture variations across natural images can thus be expressed by the empirical joint distribution of the texture parameters. After suitable post-processing, we were able to fit this distribution with a multivariate Gaussian. Sampling from the model, one obtains random vectors that can be back-transformed into texture parameters from which new textures can be synthesized. These generated textures share with natural images not only the characteristics of the second-order statistics (power spectrum) but also resemble higher-order image correlations to considerable extent. Thus, we have devised a generative model of natural texture surrogates with which we can generate fully controlled ensembles of complex stimuli useful for probing the visual system for perceptually important high-order correlations of natural images. We demonstrate the descriptive power of our model by achieving state-of-the-art compression results based on texture synthesis using a simple quantization scheme of the texture coefficients. In addition, we demonstrate how our approach can be useful for evaluating the descriptive power of generative models of natural images.

\section*{Author Summary}

Out of the total number of images that can be represented by a device (all possible pixel intensity patterns) those corresponding to photographs of the real world (``natural images") form only a tiny fraction. Similarly, in biology, where the ``device" could be a human retina and the ``pixels," photoreceptors, there is a multitude of theoretically possible receptor activation patterns, but most of them never actually occur. Visual systems may have evolved to capture the statistics of frequently occurring patterns. Therefore, an understanding of the statistical properties of {\it typical} visual input should make it possible to predict processing strategies. Finding a comprehensive mathematical description of natural images is very difficult, but even partial solutions can have wide-ranging implications, from unraveling principles underlying human perception to image compression algorithms used in DVD players or transmitting video content via the Internet. In this paper, we introduce a model of {\it natural textures}, a key constituent of natural images, representing material properties or the overall appearance of many small objects, such as leaves or grass. We show that it successfully captures the variety of textures occurring in natural images.

\section*{Introduction}

The ecological view that organisms adapt their sensory processing to the structure of their sensory input suggests a dual strategy for unravelling computational principles of perception. 
In addition to investigating the functional properties of neuronal tissue, it should be possible to derive mechanisms of sensory processing directly from statistical properties of the sensory input itself. Such mechanisms could then be related to physiological findings.
In vision, successful examples of this approach are computational models that substantiate a connection between the redundancy reduction hypothesis of Attneave and Barlow \cite{attneave_redundancy_1954, barlow_redundancy_1961} and receptive field properties of retinal and cortical neurons \cite{Atick_92, OlshausenField_96, BellSejnowski_97, vanHaterenDB, SchwartzSimoncelli_2001, Sinz2008c}.  

The commonality of such models is that they relate different elements of natural image statistics 
to certain already established neural response properties.
In order to discover hitherto unknown mechanisms of sensory processing, it is desirable to perform experiments with stimuli that have controllable image statistics and yet a complexity exceeding that of the commonly used types of noise stimuli \cite{RingachEtAl_2002, TalebiBaker_complexStimuli, VictorConte_TextureStimuli_2012, GroenEtAl_Lamme_2012_deadLeavesStimuli}.
In geneticist jargon, we need an image model with the capability to ``knock out'' certain image statistics, to study the perceptual effects of such manipulations of the stimulus statistics.

A truly profound and complete understanding of natural image statistics would include the ability to devise a comprehensive version of such a model that could capture all relevant properties of natural images. Samples generated from this model would be statistically and perceptually indistinguishable from photographs of real-world scenes.
Not surprisingly, such a model remains elusive. However, there have been substantial recent advances in addressing this problem by modeling the conditional probability distribution of the pixel intensity given the intensities within a neighborhood 
\cite{MCGSM, imageModel_Ranzato_EtAl_Hinton2013}.

\begin{figure}[!ht]
\begin{center}
\includegraphics[width=12.35cm]{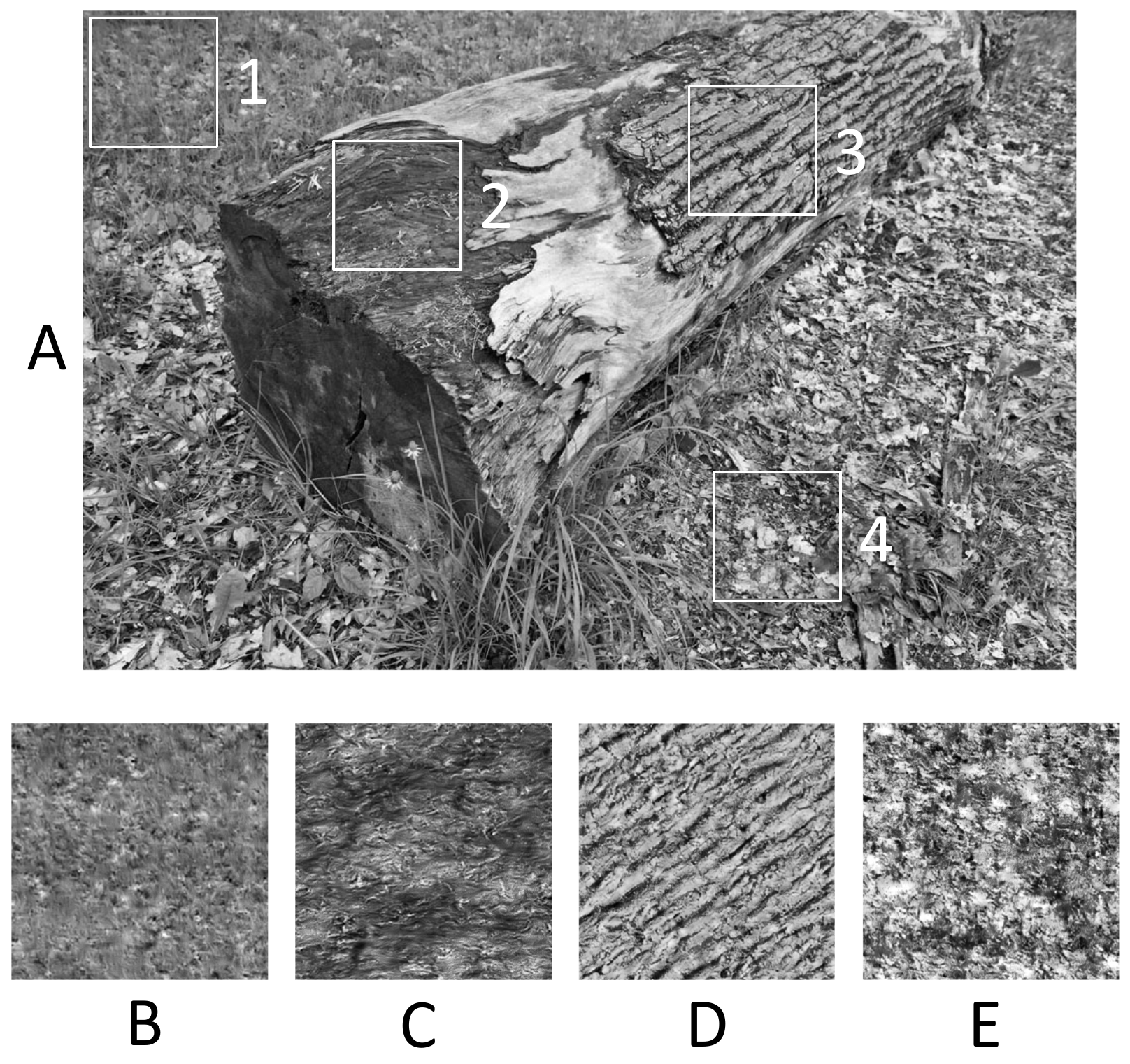}
\caption{
{\bf Natural textures and their reconstructions.}  A: natural image containing various textures. B--E: reconstructions of the textures found in the image regions 1--4 by means of the Portilla-Simoncelli algorithm \cite{PSalg}. Note, the original image patches have $128{\times}128$ pixels, whereas their reconstructions have been synthesized at $256{\times}256$ pixels. In our statistical analysis, we used even smaller patches of $64{\times}64$ pixels. It is a reasonable assumption that the texure in a randomly selected patch of this size will--in most instances--be fairly homogeneous. Moreover, we explicitly avoided transition regions between different texture regimes by means of an index of texture homogeneity introduced in equation \ref{textureHomogeneity}. }
\label{naturalTextures}
\end{center}
\end{figure}

Rather than pursuing the ultimate goal of capturing the full image complexity we focus on modeling prominent statistical characteristics of local texture variability across images, as determined by analyzing a large number of $64\times 64$ image patches which do not contain texture boundaries. The underlying philosophy is that images of natural scenes can be seen as a composition of a variety of textures (cf.~Fig.~\ref{naturalTextures}) characterizing the local fine structure within these patches, whereas  the global, large-scale structure links regions of different textures and defines the boundaries between them. Thus, we here focus on the variations in texture content and do not attempt to model the transition boundaries. 

We base our approach on the features (image statistics) employed by the parametric texture algorithm of Portilla and Simoncelli \cite{PSalg}, each of which has been shown to be perceptually prominent \cite{PSalg, balas_texture_2006}. In conjunction, these texture parameters are capable of reproducing a host of different natural textures to a significant level of detail (see Fig.~\ref{naturalTextures}).
The Portilla-Simoncelli algorithm has also been applied to modeling phenomena of peripheral vision, such as crowding and visual search, in the ``Texture Tiling Model'' of Rosenholtz and colleagues \cite{Rosenholtz_textureTilingModel}, as well as in models of extra-striate cortical processing  \cite {Simoncelli_metamers}.
 
Motivated by the high descriptive power of the Portilla-Simoncelli texture parameters, we seek to capture the natural variation of textures occurring in local regions of natural images by modeling the joint distribution of the texture parameters. More specifically, we seek to identify the empirical joint distribution of the texture parameters across a large training set of $64 \times 64$ patches from natural images. A natural starting point for density estimation is to fit a multivariate Gaussian distribution. This strategy, however, is impeded by the fact that not any combination of real numbers is a feasible parameter vector in the Portilla-Simoncelli texture model. In addition to relatively simple restrictions such as for example the nonnegativity of variance parameters, there are also more complex ones: for some combination of parameter values no texture exists that can fulfill all of the prescribed constraints simultaneously. Therefore, a major challenge of this work was to find a nonlinear re-parametrization within which for any real vector a corresponding texture can be generated. 

We were able to find such a representation and then fitted a Gaussian model to the transformed parameters. In this way it becomes straightforward to exploit a large range of tools that could not be used before. For example, we can apply principal component analysis which allows us reduce the number of required parameters by more than a factor of three. Furthermore, we can now randomly generate typical textures by drawing parameter vectors from the Gaussian distribution, remapping them into Portilla-Simoncelli parameters and synthesizing the corresponding textures. 

Borrowing terminology from nonlinear time series analysis, we refer to our sampled textures as {\em surrogates} of natural textures, in the sense that they are instantiations of a stochastic model process that shares certain properties with an imaginary ``true" process, from which the textures in actual natural scenes can be thought to originate. Even though our texture stimuli are not indistinguishable from the original texture ensemble extracted from the van Hateren database, we obtain a very good approximation and their controllable complexity makes them a promising tool for probing receptive field properties of visual neurons. 
As a critical test of the descriptive power of our density model we show that it can be turned into a lossy image compression algorithm which yields competitive performance relative to JPEG2000. Furthermore, we show how the texture parameters can also serve as a means of comparing the perceptually relevant performance of normative neural image models.

\section*{Methods}

\subsection*{Texture Analysis}
\label{texture analysis}

We take as our starting point the algorithm by Portilla and Simoncelli \cite{PSalg}, which 
employs a multi-scale image representation in the form of a complex-valued steerable filter pyramid \cite{steerPyr,steerableFiltersSimoncelli,steerableFilters}. The real and imaginary parts of the complex filter coefficients are the responses of even and odd-symmetric oriented filters. In addition to accessing individual filter responses, one can back-transform the combined contributions of all filter responses within a given scale (pyramid level) and obtain a {\it partially reconstructed image}, which is simply a bandpass-filtered version of the input image in the corresponding frequency range.

By evaluating certain statistics of pixel intensities and oriented filter responses, the algorithm obtains a set of texture parameters that model the image structure (texture analysis). The particular statistics considered are:

\begin{itemize}
\item{Marginal pixel statistics, i.e. mean, variance, skewness and kurtosis of pixel intensity;}

\item{Skewness and kurtosis of the partially reconstructed image at each scale (including the lowpass residual);} 

\item{Central autocovariances of the partially reconstructed image at each scale (including the lowpass residual) within an $N_a{\times}N_a$ pixel neighborhood, where typically $N_a=7$; this being a compact way of representing the power spectrum;}

\item{Central autocovariances of filter coefficient magnitudes at each scale and for each orientation, within an $N_a{\times}N_a$ pixel neighborhood;}

\item{Covariance matrix of filter coefficient magnitudes across orientations within each scale;}

\item{Cross-covariance matrix of filter coefficient magnitudes across orientations at adjacent scales;}

\item{Cross-scale phase statistics, measured by the cross-covariance matrix of complex filter coefficients at adjacent scales (real with real parts and real with imaginary parts).}

\end{itemize}

Each of the above listed parameter groups has been shown to be perceptually relevant for texture synthesis \cite{PSalg}, but they do \emph{not} form a  complete set of constraints, i.e.~one that can fully capture the perceptually assessable characteristics of any given texture. Typically, the number of parameters required for an accurate representation of a texture is of the order of several hundred. For instance, employing a filter pyramid with four scales, four orientations, and  autocovariances evaluated within a neighborhood of 7$\times$7 pixels results in a set of 655 parameters. 
As a result, the input image is transformed into a compact, probabilistic representation requiring at least an order of magnitude fewer paramaters than there are pixels. We next describe the inverse of this transformation: texture synthesis.

\subsection*{Texture Synthesis}
 
Texture synthesis is a process whereby one can obtain a reconstruction of the corresponding texture from a given set of parameters. The parameters play the role of statistical constraints to be enforced. 
 Commencing with an initial random image, this is accomplished via a gradient descent procedure that iteratively adjusts pixel intensities and wavelet coefficients until their marginal and pairwise statistics have converged sufficiently to their desired values. Repeating the same procedure with other random initializations produces new instantiations of the same texture. For details, the reader is referred to \cite{PSalg}.


\subsection*{The Meta-statistics of Natural Images}
\label{metastatistics}

The Portilla-Simoncelli algorithm \cite{PSalg} provides a model of a texture in the form of a parameter vector. Such a vector can be thought of as referencing a point in the high-dimensional abstract space of all representable textures, given the set of statistics employed by the algorithm. 
The parameter vectors corresponding to natural textures live in a particular subspace, and in understanding the structure of this space, one can gain insight into natural image statistics in terms of a model of the distribution of texture parameters across an image ensemble. Since the parameters are themselves statistical features (of individual images), we refer to their joint distribution across an image ensemble as the \emph{meta-statistics}.

One way of obtaining an approximate, quantitative description of the \emph{meta-statistics of natural textures} is to fit a parametric model to the empirical distribution of texture parameter values over a large set of randomly selected patches extracted from a database of natural images. 
The texture characteristics in photographs of natural scenes tend to be highly variable due to the presence of different kinds of overlapping objects, such as foliage, branches, and twigs.
Therefore, we consider small patches wherein the texture is fairly homogeneous (cf.~Fig.~\ref{naturalTextures}) and study the variability across a large ensemble. In addition, we introduce a criterion to exclude transitional regions wherein the texture characteristics change abruptly within a patch. 

The degree of texture homogeneity $h$ within an image patch is determined by dividing it into four sub-patches and evaluating the variability of the sub-patch power spectra $|F_j(\omega)|^2$ around the mean spectrum $\langle | F(\omega)|^2 \rangle$, across all $N$ discrete frequencies $\omega_i$:

\begin{equation}
h = \sum_{i=1}^{N} \sum_{j=1}^{4} \log \frac{ \langle | F(\omega_i)|^2 \rangle } { |F_j(\omega_i)|^2 }.
\label{textureHomogeneity}
\end{equation}
This measure is motivated by the plausible underlying assumption that variations in the local power spectrum very likely coincide with changes in higher-order image correlations and, conversely, that a roughly constant power spectrum is also an indicator for homogeneous higher-order correlations across a patch.
Sorting a randomly selected set of patches according to texture homogeneity in descending order and discarding the lower half of the set helps to evaluate the meta-statistics over a more suitable set of training data.

Once fitted to the natural image data, a parametric meta-model provides a joint probability density from which new texture parameter vectors can be sampled. The corresponding textures may be quite unlike any of the training examples, but they will be surrogates of natural textures, since the joint distribution of their Portilla-Simoncelli texture parameters is, to some extent, matched to that of natural image patches. 
In the following section, we introduce an example of a meta-model in the form of a multivariate Gaussian.

\subsubsection*{A multivariate Gaussian model}
\label{Gaussian model}

Ideally, one would wish to capture the full joint statistics of the texture parameters, but this is, of course, impractical due to the dimensionality of the parameter space. 
Therefore, we restrict our approach to the simplest model able to approximate the pairwise joint statistics of all texture parameters, a multivariate Gaussian with an $n \times n$ covariance matrix $\Sigma$, where $n$ is the number of parameters. Despite its simplicity, the Gaussian meta-model is not a trivial one, since many of the texture parameters are nonlinear features capturing higher-order image correlations.

Since the marginal distributions of the parameters are typically highly non-Gaussian, we first apply different invertible transforms described below, in order to make the marginal distributions roughly symmetrical and approximately normal. The particular transformation applied depends on the type of texture parameter, the choice being heuristic and motivated by the observed Gaussianizing effect on the empirical marginal distribution.
Consequently, the second-order joint statistics across the image ensemble thus obtained will be that of the \emph{transformed} texture parameter vectors. 
In addition, we take into account some \emph{a priori} known constraints among parameters, since at the subsequent sampling stage these constrains must be fulfilled in order to obtain valid parameter sets.
 
We now describe the transformations applied to the texture parameter values.
Positive quantities, such as variances, are log-transformed. Autocovariances are first normalized by their central variance (the value of the autocovariance function at lag zero) and then approximately symmetrized by means of the inverse hyperbolic tangent. Analogously, cross-covariances are normalized into cross-correlations and then likewise transformed. The inverse hyperbolic tangent maps the range of correlation values from the finite interval $[-1,1]$ to the real line, such that their distribution can be approximated by a Gaussian. 
In order to symmetrize the distribution of skewnesses, we apply the \emph{modulus transform}  \cite{modulusTrafo}, defined as:

\begin{equation}
y \mapsto
 \begin{cases}

 \text{sign}  (y)  \left[   \displaystyle \frac{ (|y| + 1)^\lambda } { \lambda}  \right], & \lambda \ne 0, \\[2ex]
  \text{sign}  (y)  \left[ \log(|y|  + 1) \right], & \lambda = 0,

\end{cases}
\end{equation}
\vspace{1ex}

\noindent
with a manually fitted parameter $\lambda = 0.1$.

Filter response covariance matrices are normalized to correlation coefficient matrices and Cholesky-factorized. Their meta-statistics are evaluated using the entries of the Cholesky matrices, which carry a negative or positive sign and are roughly normally distributed.
At the sampling stage, the process is reversed by interpreting the corresponding sampled parameter values accordingly: as entries of a triangular matrix $M$ from which a positive definite matrix can be obtained by computing $C=MM^T$. Thereby we ensure that the sampled matrix $C$ is a valid filter response covariance matrix.

 The skewness $s$ of a distribution sets a lower bound for the corresponding value of the kurtosis $\kappa$ \cite{skewnessBound}: 
\begin{equation}
\label{kurtosisBound}
 s^2  + 1  \le  \kappa ,
 \end{equation}
where the kurtosis is defined such that $\kappa=3$ in the Gaussian case. 
We incorporate this relationship by defining the quantity
\begin{equation}
\tilde{\kappa} = \kappa - s^2 - 1.
\end{equation}
Since $\tilde{\kappa}\ge 0$ and in practice $\tilde{\kappa}$ is actually positive, we choose $\log \tilde{\kappa}$ as our transformed kurtosis parameter, which fulfills the constraint in Equation \ref{kurtosisBound}.
Having introduced the necessary transformations to obtain approximately Gaussian marginal parameter distributions,
fitting a multivariate Gaussian to the transformed texture parameter data seems straightforward. However, when drawing texture parameters from this meta-model, the texture synthesis usually does not converge. This is the result of a lack of intrinsic consistency among the generated texture parameters, which we address in the following section. 
 
 
\subsubsection*{Consistency between second-order and higher-order image statistics}


For our purpose of fitting a meta-model and generating new textures from this model, the procedures for image analysis and synthesis in the Portilla-Simoncelli algorithm are not readily applicable. 
The underlying program code is designed to analyze and subsequently resynthesize a given texture. In that case, the parameters supplied to the synthesis routine can be considered {\em valid by default}, since the entire set of values was obtained from an actual image. Even though there is no guarantee that the resynthesis process actually converges, empirically this seems to be the case for the overwhelming majority of textures. 
The intuitive explanation is that the values of different parameter groups originating from the same image are necessarily consistent with one another, which makes it possible to iteratively enforce the corresponding image statistics. 
In order to achieve the same intrinsic consistency in parameters sampled from our model, which is based solely on \emph{pairwise} statistics, we must ensure that the parameters satisfy additional constraints, described in the following.

\begin{figure}[!ht]
\begin{center}
\includegraphics[width=8.3cm]{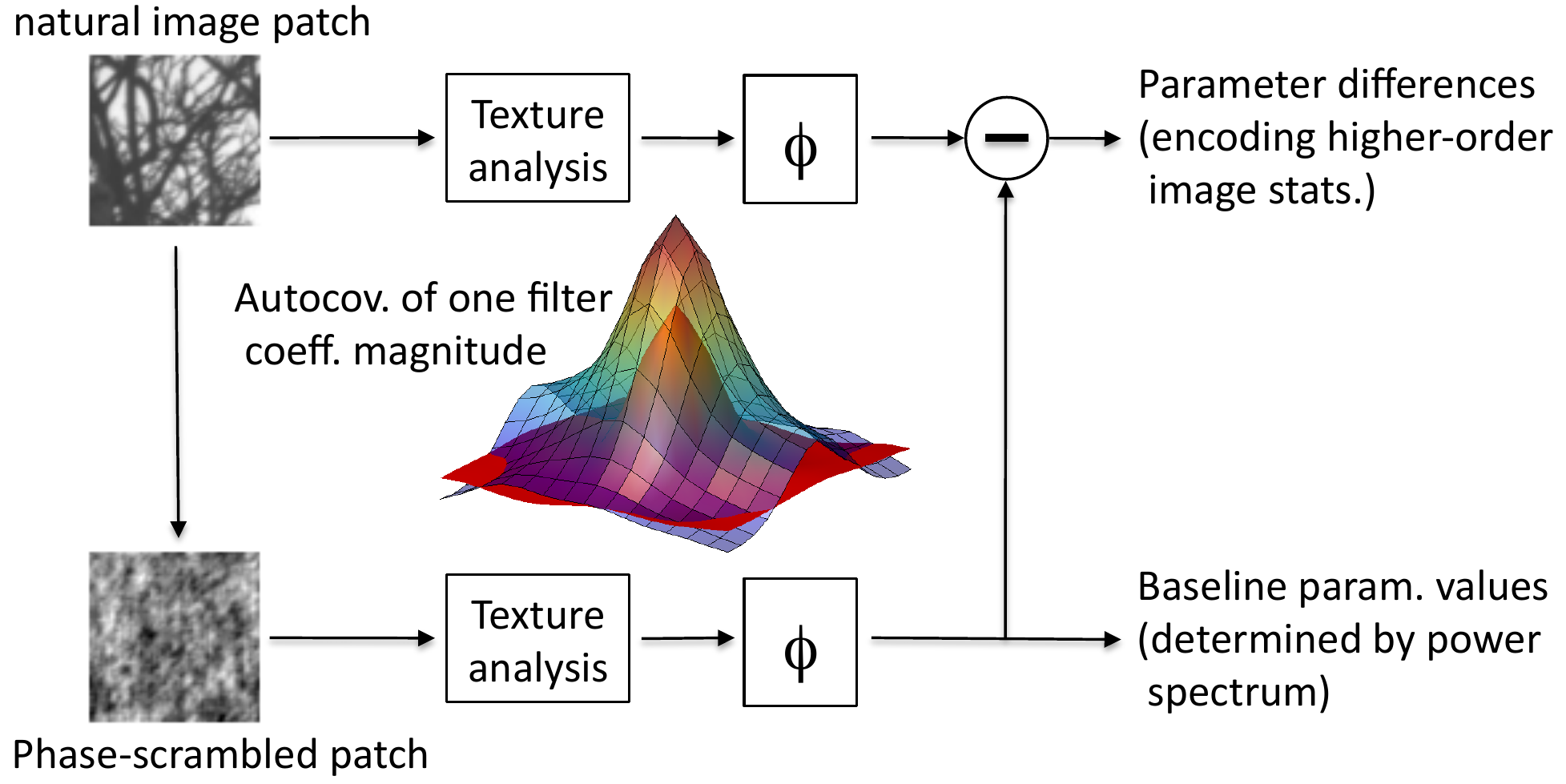}
\caption{
{\bf Schematic diagram of our texture analysis procedure.} An image is first analyzed using the Portilla-Simoncelli texture analysis algorithm. In addition, a phase-scrambled version of the image is analyzed in the same manner, yielding the ``baseline" values of the parameters. Both texture parameter vectors undergo symmetrizing transformations, represented by the map $\phi$, and the baseline values are subtracted from the components of the original parameter vector. 
An example of baseline vs. original parameters is given by the 3D plots of a magnitude autocovariance function for one particular orientation and scale. The red surface shows the residual baseline magnitude autocovariances that persist even when higher-order image correlations have been removed by phase scrambling. The difference between the surface plots (the ``net autocovariance function'') is used for meta-statistical analysis.}
\label{analysisScheme}
\end{center}
\end{figure}

In analyzing the Portilla-Simoncelli texture parameters, we found a strong dependence between parameters encoding second-order image correlations (SOC-parameters) and those encoding higher-order image correlations (HOC-parameters).
Therefore, it is necessary to lay the groundwork for consistency among these two parameter groups early on, as part of a modified texture analysis procedure (Fig.~\ref{analysisScheme}).
In addition to the original input image, we also texture-analyze a phase-scrambled version of the image, meaning that the phases of its complex Fourier coefficients have been randomized such that phase correlations and phase-magnitude correlations are destroyed.
Even with the higher-order image correlations removed, the HOC-parameters assume non-trivial values, unless the input image is virtually structureless (Gaussian white noise). For instance, the spatial autocovariance functions of the filter coefficient magnitudes obtained from a phase-scrambled natural image are not delta functions, though they do decay more rapidly than the autocovariance functions of the original image (see the inset graph in Fig.~\ref{analysisScheme}).  
For a phase-scrambled image, the HOC-parameter values are redundant, in the sense that they are fully determined by the power spectrum (i.e.~the  SOC-parameters). We will refer to these as \emph{baseline HOC-parameter values}.
 
When higher-order correlations \emph{are} present, the actual values of the HOC-parameters differ from their theoretical baseline values, but not arbitrarily. For instance, in the above-mentioned example of the autocovariance function, the decay of the baseline autocovariance function sets a lower bound for the decay of the actual autocovariance function.
Thus, the underlying dependency between HOC and SOC-parameters is one of restrictions that the SOC-parameters place on the possible HOC-parameter values. Sampled texture parameters must also fulfill these constraints.
Otherwise, the lack of intrinsic consistency can introduce spurious higher-order correlations into the textures drawn from the model or result in non-convergent synthesis. 

We solve the consistency problem by expressing all HOC-parameters relative to their baseline values.
For example, all parameter groups representing spatial autocovariance functions are normalized to autocorrelations. Only the center elements of autocovariance functions (i.e.~the variances) are taken in their original (log-transformed) form, since they serve as  a record of the absolute scale of the original autocovariance functions. 
We then subtract the baseline HOC-parameter values from their original values and operate with the differences. 
With the redundant part subtracted, the modified HOC-parameters and the SOC-parameters are effectively disentangled,  
and one can proceed to evaluate the pairwise statistics of all parameters across natural images.

\subsubsection*{Estimating the hyperparameters}
\label{EstimatingHyperparameters}

In order to obtain reliable estimates of the hyperparameters of the multivariate Gaussian model (the mean parameter vector $\mu$ and the parameter covariance matrix $\Sigma$), the number of image patches should be at least of the order of the number of non-redundant entries in the parameter covariance matrix.
Therefore, we extracted $\sim 10^6$ patches of 64$\times$64 pixels at random locations from 3600 images in the van Hateren database \cite{vanHaterenDB}, avoiding transitional regions using our homogeneity measure defined in (\ref{textureHomogeneity}).
The choice of patch size constitutes a trade-off between minimizing the patch size (thus maximizing texture homogeneity) and maximizing the amount of data (pixels) available to evaluate the image statistics (texture parameters). Empirical tests showed that a patch size of 64$\times$64 pixels yields a good compromise. 


Let $\phi$ be a vector-valued function comprising all Gaussianizing transformations to be applied to the components of a texture parameter vector $\bf{f}$, as well as the normalizing of covariances to correlations and Cholesky decompositions of orientation covariance matrices, as introduced previously. Then the sample covariance matrix of the transformed parameters is calculated as 
\begin{equation}
\Sigma = E\left \{ \left[ \phi(\mathbf{f}) - \mu \right]  
				    \left[ \phi(\mathbf{f}) - \mu \right]^T \right \},
\label{sampleCovMatrix}
\end{equation}
\noindent
where $E\{.\}$ denotes the expectation computed over the entire set of training images, and  with a mean parameter vector $\mu =  E\{ \phi(\mathbf{f}) \}$.



 \subsection*{Generating Textures from Sampled Parameters}

Having estimated the hyperparameters that characterize the Gaussian model of the joint density of texture parameters, one can sample from this density to obtain new parameter sets. 
First, we diagonalize the empirical covariance matrix $\Sigma$ via singular value decomposition:
\begin{equation}
\Sigma = U D V^T.
\label{diagonalization}
\end{equation}
\noindent
A root of $\Sigma$ is then given by $U \sqrt{D} $,  which we use to transform a Gaussian random vector $\mathbf{x}$, with independently drawn zero-mean and unit-variance components, into a new vector with components obeying the second-order statistics defined by $\Sigma$ and the mean vector $\mu$:
\begin{equation}
\mathbf{f}^{(s)} =  U \sqrt{D}  \: \mathbf{x} + \mu. 
\label{rawParameterSample}
\end{equation}

However, $\phi^{-1} (\mathbf{f}^{(s)} )$, as obtained from applying the inverse Gaussianizing transform
to the result of equation (\ref{rawParameterSample}), does not constitute a texture parameter vector compatible with the Portilla-Simoncelli texture synthesis routine, for two reasons. First, the components corresponding to HOC-parameters are to be interpreted as difference-from-baseline values, to which the yet to be determined baseline values must be added to construct the actual HOC-parameter values. Second, we found that the ``raw'' sampled values of the SOC-parameters lack cross-scale consistency, leading to non-convergent texture synthesis.
In the next subsection, we introduce a modified texture synthesis procedure that reconstitutes the HOC-parameters
and resolves the cross-scale inconsistencies of SOC-parameters.

\subsubsection*{Ensuring cross-scale and intrinsic consistency}
 \label{scale_consistency}
 
 Our modified texture synthesis algorithm consists of three steps. First, from the sampled SOC-parameters we obtain a cross-scale consistent set of SOC-parameters. Second, using this new set of SOC-parameters, we derive the baseline values of the HOC-parameters. Third, we add the baseline values to the sampled difference-to-baseline HOC-parameters to construct the actual HOC-parameters.
 
In order to achieve cross-scale consistency of the SOC-parameters, we make use of the additivity property of the steerable pyramid image representation. This allows us to enforce the second-order image statistics separately at each scale, where convergence is easily reached. 
By separately inverting the steerable filter transform, we obtain a partially reconstructed image from each scale. Summing up the partial reconstructions yields a new image with a power spectrum equal to that of the new texture, but no higher-order correlations. Performing a regular Portilla-Simoncelli texture analysis on the summed image results in a full-length texture parameter vector $\mathbf{f}^{(0)}$ containing a new set of SOC-parameters that is now necessarily scale-consistent and that replaces the initial SOC-parameter samples. 
Symbolically, we write this parameter vector as a concatenation of three vectors representing the three basic parameter groups (marginal pixels statistics, second-order image correlations and higher-order correlations):

\begin{equation}
\mathbf{f}^{(0)} = \left[ \mathbf{f}^{(0)}_{\mbox{\footnotesize marg.}}, \mathbf{f}^{(0)}_{\mbox{\footnotesize SOC}},  \mathbf{f}^{(0)}_{\mbox{\footnotesize HOC}}  \right] .
\end{equation}
\noindent
Moreover, since the summed image contains only second-order image correlations, the components of $\mathbf{f}^{(0)}_{\mbox{\footnotesize HOC}}$ are the very baseline values required to construct the HOC-parameters. 
Written in the same symbolic manner, the structure of the sampled parameter vector is:
\begin{equation}
\mathbf{f}^{(s)} = \left[ \mathbf{f}^{(s)}_{\mbox{\footnotesize marg.}},\mathbf{f}^{(s)}_{\mbox{\footnotesize SOC}},  \mathbf{\Delta}^{(s)}_{\mbox{\footnotesize HOC}} \right ] ,
\end{equation}
\noindent
where the symbol $\mathbf{\Delta}^{(0)}_{\mbox{\footnotesize HOC}}$ indicates that the sampled HOC-parameters are to be interpreted as difference-to-baseline values. Thus, all the necessary components for constructing a valid parameter vector are at hand. Denoting this new vector by $\mathbf{f}^*$, it can be written as: 

\begin{equation}
\mathbf{f}^* = \phi^{-1} \left[ \mathbf{f}^{(s)}_{\mbox{\footnotesize marg.}},  
							\phi \left( \mathbf{f}^{(0)}_{\mbox{\footnotesize SOC}} \right),  
			                         \phi \left( \mathbf{f}^{(0)}_{\mbox{\footnotesize HOC}} \right) + 
			                          					\mathbf{\Delta}^{(s)}_{\mbox{\footnotesize HOC}} \right ].
\end{equation}
\noindent
Here, the parameters encoding marginal statistics are exactly as sampled; the SOC-parameters are taken from the above described baseline vector vector $\mathbf{f}^{(0)}$, since they are cross-scale consistent by construction; and the HOC-parameters are obtained by adding baseline values to the sampled difference-to-baseline values. The transformation $\phi$ is applied to the components of the baseline vector, whereas the sampled values are to be interpreted as already transformed, since the Gaussian meta-model was fitted to transformed parameter values (cf. equation \ref{sampleCovMatrix} ).
The inverse mapping $\phi^{-1}$ involves applying the inverse of the Gaussianizing parameter transforms but also the reconstitution of autocovariance functions from autocorrelations, as well as fulfilling the previously mentioned constraints, such as the positive definiteness of orientation covariance matrices or the relationship between skewness and kurtosis parameters. 

After this procedure, the resulting parameter vector $\mathbf{f}^*$ is compatible with Portilla and Simoncelli's original texture synthesis routine.

\section*{Results}

We have devised a generative model of naturalistic textures (natural texture surrogates) based on the statistical features (texture parameters) employed by the Portilla-Simoncelli (PS) algorithm \cite{PSalg}. 
The PS algorithm represents an image by a steerable filter pyramid, a multi-resolution image representation by means of the responses of oriented filters (wavelets), which are biologically motivated since they resemble receptive field properties in the mammalian primary visual cortex. 
The texture parameters capture the intensity histogram via its first four moments (mean, variance, skewness and kurtosis), the power spectrum in the form of spatial autocovariances, and a substantial amount of higher-order image correlations by means of the variances and (auto)covariances of the oriented filter responses. The excellent texture reconstruction capability of the PS algorithm (see Fig.~\ref {naturalTextures}) demonstrates that its parameters can capture the perceptual characteristics of many natural textures, or at least important elements thereof.

Building on the Portilla-Simoncelli methodology for texture analysis and synthesis, we explored the feasibility of devising a texture generator by fitting a multivariate Gaussian to the empirical joint distribution of the PS texture parameters across small patches (64$\times$64 pixels) from a large ensemble of natural images. As a repository of natural images, we used the van Hateren image database \cite{vanHaterenDB}. 
However, sampling from the Gaussian model, we obtained parameters with which the texture synthesis procedure would not converge. Therefore, we had to substantially modify the original texture analysis and synthesis procedures devised by Portilla and Simoncelli, in order to endow our sampled texture parameters with the intrinsic consistency required for successful texture synthesis (see the Methods section for details).

\begin{figure}[!ht]
\begin{center}
\includegraphics[width=17.35cm]{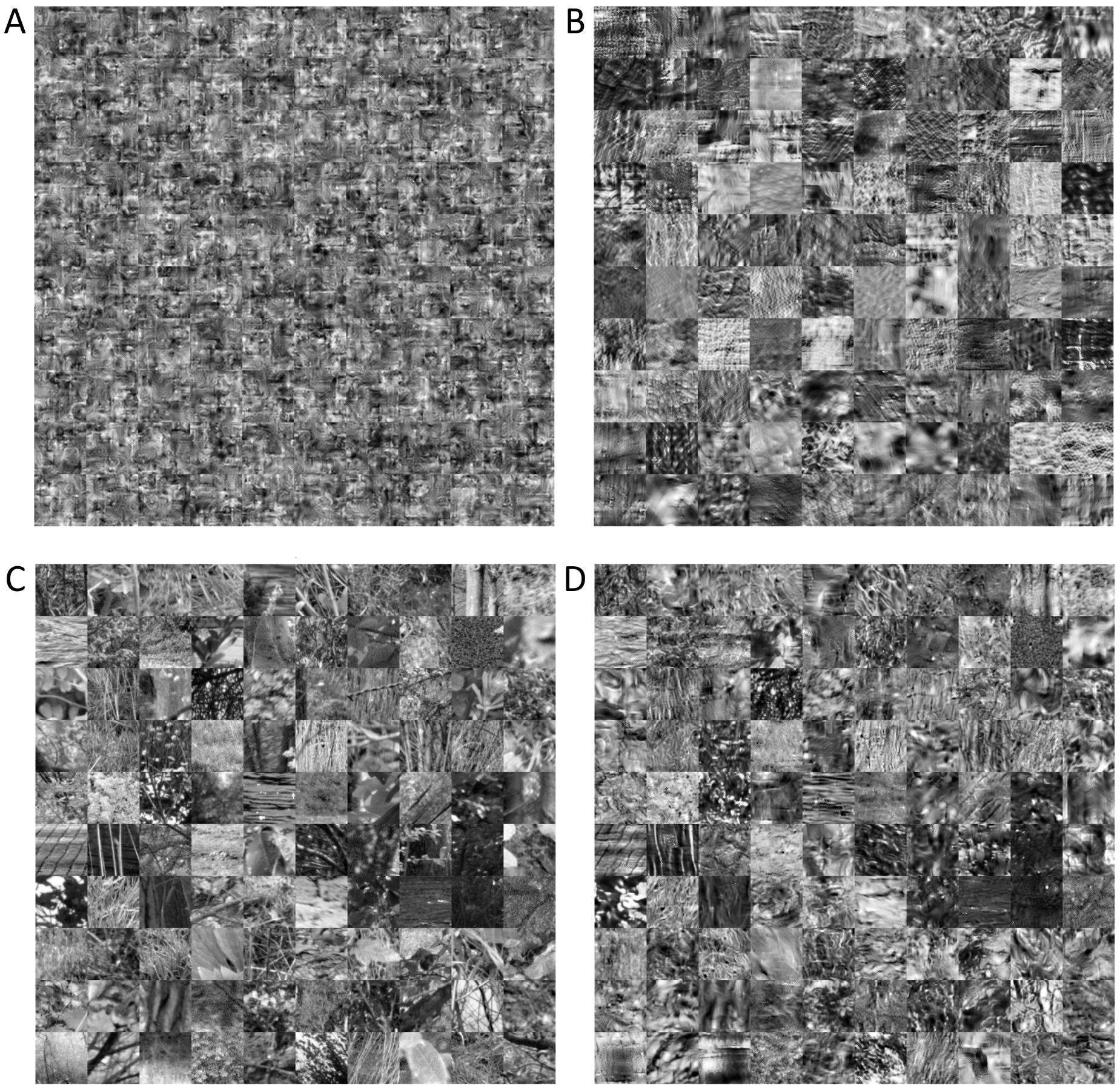}
\caption{
{\bf Natural and artificial texture patches.} 
 A: 100 instantiations of the mean texture of our meta-model trained on natural image patches ($64{\times}64$ pixels). B: 100 textures drawn from the meta-model using both mean vector and covariance matrix. C: 100 natural patches from the training set (van Hateren image database). D: the same patches as in C, resynthesized using our modified versions of the Portilla-Simoncelli texture analysis and synthesis algorithms.
 The comparison of generated (B) and resynthesized textures (D) reveals the existence of higher-order parameter correlations not captured by our multivariate Gaussian model. While being more naturalistic than our model samples, the resynthesized natural patches (D) do not fully capture the statistics of the originals (C). In particular, sharp long-ranging contours are difficult to reproduce.}
\label{ExamplesTextures}
\end{center}
\end{figure}

Figure \ref{ExamplesTextures}A shows 100 example textures sampled from our model using only the mean parameter
vector across natural image patches. These instantiations of the ``averaged natural texture" (each synthesized starting with a different noise initialization image) are perceptually quite remote from any naturalistic texture. However, the texture samples generated with our full model (mean parameter vector plus covariance matrix), shown in Figure \ref{ExamplesTextures}B, have a more ``organic" appearance.
For comparison, we also provide 100 natural patches (Fig.~\ref{ExamplesTextures}C) from the van Hateren image database, similar to those in the training set, as well as the patches' reconstructions by means of the PS algorithm (Fig.~\ref{ExamplesTextures}D). 
The comparison of panels C and D shows certain limitations of the PS algorithm in modeling natural patches, whereas comparing panels B and D reveals limitations of our Gaussian meta-model to correctly reproduce the frequencies with which the different textures occur in natural images.


\subsection*{Natural Texture Surrogates as Complex Stimuli}

Having a generative model for natural texture surrogates can be useful in various ways. An important application for psychophysical or neurophysiological experiments is the possibility to construct textures which form a middle ground between the classical grating-type stimuli and actual natural images, in terms of the control vs.~complexity trade-off \cite{FreemanEtAl_V2signature, VictorConte_TextureStimuli_2012, TalebiBaker_complexStimuli}.

The example textures shown in Figure \ref{ExamplesTextures} have 64$\times$64 pixels, which is the same size as the 
training patches used to fit our model.
Alternatively, the sampled textures can be synthesized in a larger format (limited in practice only by the amount of memory required for the filter pyramid representation of the image), which is a basic feature of the Portilla-Simoncelli algorithm, since the texture parameters are merely statistical descriptors with reference to spatial frequency, not absolute image size. 
The training patch size merely sets a lower bound for the spatial frequency range (number of pyramid levels) within which image correlations can be enforced. For instance, a patch with 64$\times$64 pixels cannot be represented by more than $\log_2(64)=6$ scales (five-fold downsampling)

In practice, four spatial scales have been found to suffice. In such a filter pyramid, the coarsest scale of a 64$\times$64 patch is only represented by 8$\times$8 = 64 coefficients (three-fold downsampling), leaving only 4$\times$4 = 16 coefficients for the low-pass residual.
Hence, representing a fifth scale (via a further downsampling step) would add very little information, as far as filter response correlations are concerned, and would make it difficult to obtain reliable estimates of marginal statistics, such as the pixel skewness and kurtosis, of the coarsest frequency band.

When stimuli are synthesized in larger format than 64$\times$64 pixels, texture synthesis is still based on the number of scales used for analysis, i.e.~a four-scale filter pyramid, which contains fewer spatial scales than theoretically possible given the larger output image size. However, since the information is not available, no image correlations in the frequency range below the fourth scale can be enforced. 
As a result, the stimuli are mostly structureless in the lowest spatial frequency range, but, unlike with high frequency noise, there are no obvious visual artifacts. 

The Stimulus structure is well-defined and controllable, since the experimenter can influence the image correlations by choosing which image statistics to use for texture synthesis. 
There are two ways of controlling the image statistics of the stimuli. First, during synthesis, different parameter groups can be included or excluded, which provides a means of switching the corresponding image correlations on or off. Manipulating \emph{individual} parameters is generally problematic due to the strong parameter interdependencies, which may make texture synthesis impossible.
Second, one can diagonalize the parameter covariance matrix and operate in the eigenbasis of the parameter space, where any linear combination of eigenvectors corresponds to a valid parameter vector that can be used for texture synthesis. 
The advantage of the first method is the meaningfulness of the different parameter groups, whereas the principal components are somewhat abstract, since there is no intuitive association between individual eigenvectors and particular image statistics. 
However, by definition, each coefficient in the eigenbasis can be varied independently, which is useful for system identification experiments.

\subsubsection*{Reverse correlation experiments with texture stimuli}

Even though our meta-model cannot capture all the statistical regularities of natural image patches,
the sampled textures are ideally suited to serve as a form of complex colored noise in system identification experiments.
In fact, our stimuli have a higher complexity and closer resemblance to natural images than the artificial stimuli employed in sophisticated system identification experiments \cite{RingachEtAl_2002, DavidVinjeGallant_2004, TalebiBaker_complexStimuli}. 
In addition to using white noise or sequences of bars or gratings, the authors of these studies demonstrate the benefits of naturalistic stimuli for evoking more pronounced neural responses than otherwise possible, as well as for the activation of nonlinear processing in V1 and extra-striate cortical areas. However, this approach comes at the price of having virtually no control over stimulus statistics.  
Our natural texture surrogates can provide a middle ground between ``classical" artificial and natural stimuli, \emph{without} compromising control.  
 The stimuli are generated in the eigenbasis of the PS texture parameter space by drawing ``white" random vectors (the components have independent Gaussian distributions), transforming them into the regular PS parameter space, and synthesizing the corresponding textures. Thus, one can obtain sequences of arbitrarily many stimuli, record the elicited responses of visual neurons, and determine the receptive field structure via system identification. 

In the following section, we analyze the parameter covariance matrix and investigate the implications of our approach for understanding the structure of natural images and for possible applications resulting from this knowledge.


\subsection*{The Principal Components of the Texture Parameter Space }

\begin{figure}[!ht]
\begin{center}
\includegraphics[width=8.3cm]{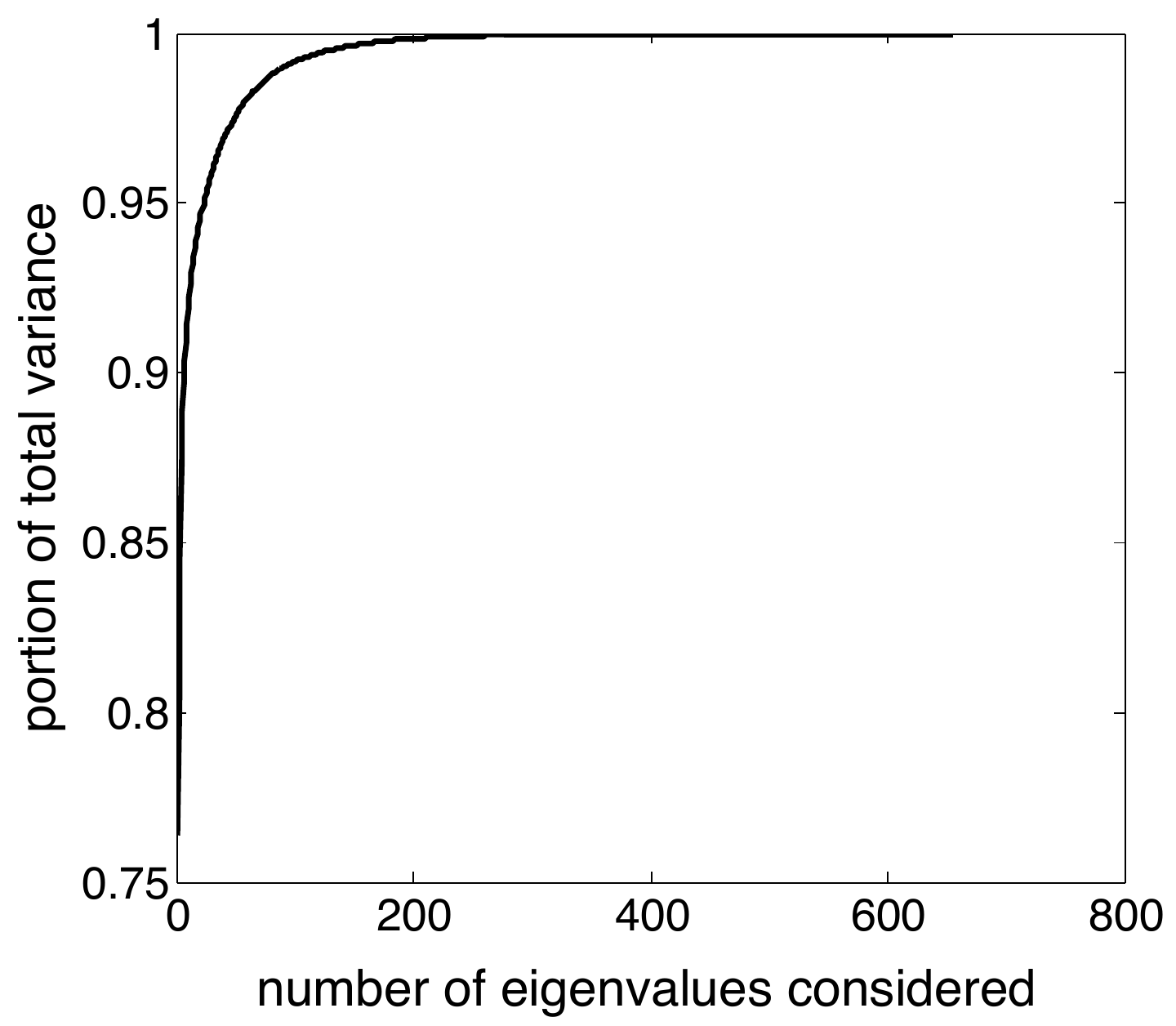}
\caption{
{\bf Principal component analysis of the parameter covariance matrix.} The covariance matrix of Portilla-Simoncelli texture parameters was obtained from a large ensemble of natural image patches ($N>10^6)$. A plot of the portion of total variance explained against the number of considered eigenvalues indicates substantial redundancy. About 200 eigenvectors should be sufficient to represent natural textures with reasonable accuracy.}
\label{cumVariance}
\end{center}
\end{figure}

Having diagonalized the parameter covariance matrix obtained from natural image patches for the purpose of generating texture samples, one can determine whether there is redundancy among the parameters that permits a dimensionality reduction in the eigenspace. We shall refer to the set of eigenvectors as the \emph{natural eigenbasis}. 
Redundancy can be visualized by plotting the portion of total variance explained (i.e.~the cumulative sum of the eigenvalues divided by their sum) as a function of the number of eigenvectors taken into account 
(Fig.~\ref{cumVariance}).
Provided that the natural eigenbasis has been obtained from a \emph{representative} natural image ensemble, retaining only about 200 dimensions should suffice for encoding (and reconstructing) any natural texture.

\begin{figure}[htbp]
\begin{center}
  \includegraphics[width=17.35cm]{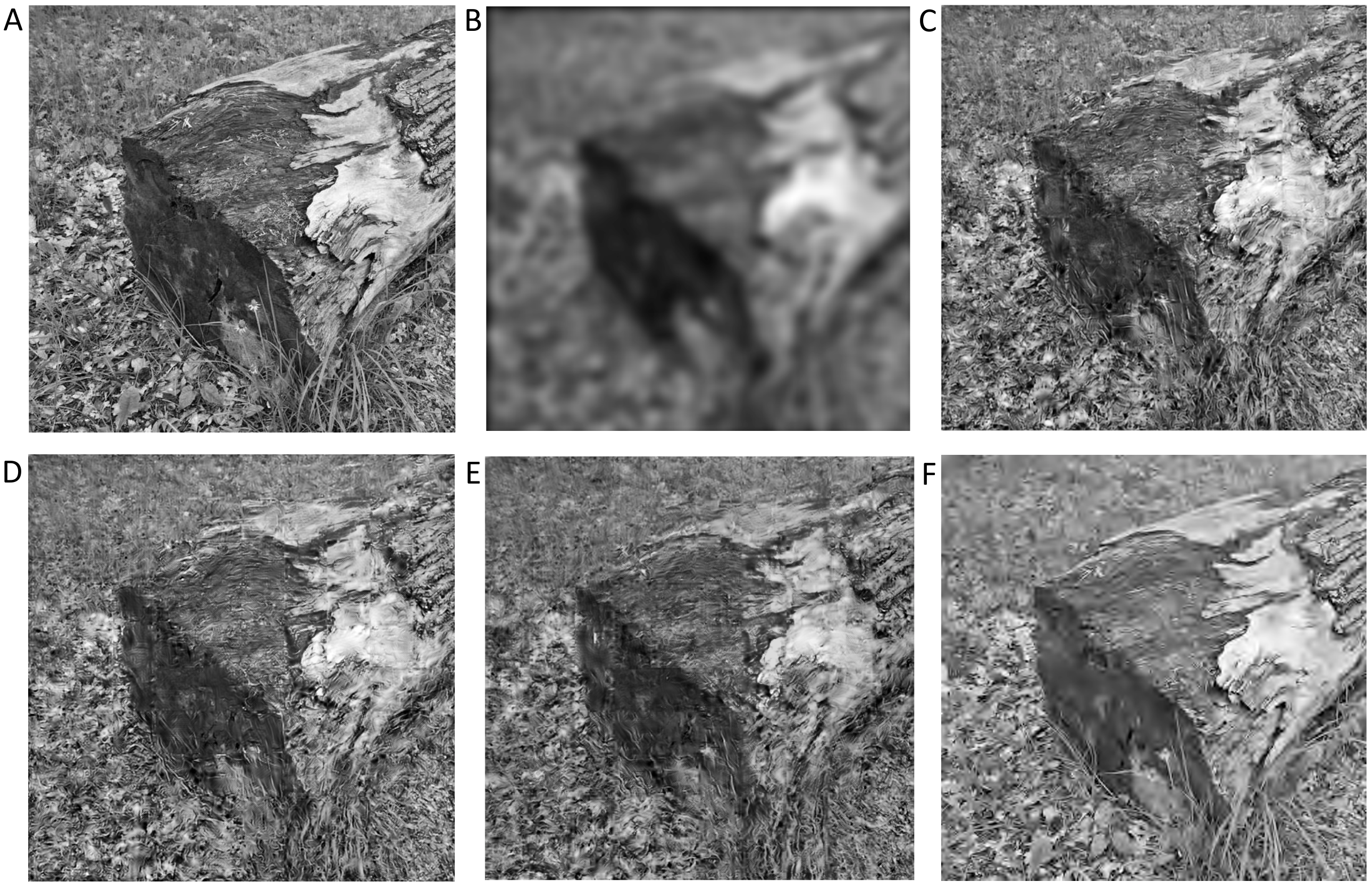}
\caption{
{\bf Approximate texture reconstruction in a natural image.} A: original image; B: low-frequency part of the image used as an initialization for texture synthesis; C: patch-by-patch reconstruction of A (10$\times$10 patches, each with 64$\times$64 pixels), using the low-pass filtered image (B) as an initialization for texture resynthesis; D: same as C but using a dimensionality-reduced representation of the parameter vectors in each patch, restricted to the 200 most significant eigenvectors. As expected based on Fig.~\ref{cumVariance}, this dimensionality reduction introduces very little degradation in the quality of image reconstruction. E: using again 200 eigenvectors but with quantized eigenbasis coefficients and low-frequency initialization (DCT-compressed version of B), encoded at 0.1376 bits/pixel. F: equivalent JPEG2000 version of A (encoded at 0.1389 bits/pixel).}
\label{dimensionalityReduction}
\end{center}
\end{figure}

To test this prediction we divided a 630$\times$630 pixel natural image (Fig.~\ref{dimensionalityReduction}A)
into 100 patches, each having 64$\times$64 pixels and overlapping by one pixel, and attempted a patch-by-patch reconstruction of the image by analyzing and resynthesizing the texture in each patch. To reduce blocking artifacts, intensities in the overlapping regions were interpolated after each iteration during texture synthesis. In order to achieve global consistency of patches, we used a low-pass filtered version of the image (Fig.~\ref{dimensionalityReduction}B) as an initialization for texture synthesis (a technique previously used in \cite{Balas_lowpass}). Guided by the low-spatial frequency information, the patch-by-patch synthesis yields an image (Fig.\,\ref{dimensionalityReduction}C) quite close to the original, apart from some inaccuracies in the reproduction of long-ranging sharp edges and blocking artifacts.

In a second step, we repeated the above procedure, but after texture analysis the parameter vector of each patch was transformed into the natural eigenbasis and all components above dimension 200 were set to zero. After back-transforming and resynthesis, we obtained the image corresponding to this dimensionality-reduced representation (Fig.\,\ref{dimensionalityReduction}D). By comparing C and D, one can appreciate only a mild degradation in the quality of reconstruction, confirming that the subspace of the dominant 200 eigenvectors does indeed capture most of the perceptually relevant information about local texture in the image.
This observation is even more significant considering that none of the reconstructed patches are part of the training set used to obtain the natural eigenbasis. 
The success of the dimensionality reduction experiment confirms the general validity of the natural eigenbasis obtained from the van Hateren image set. 


\subsubsection*{Perceptual importance of natural eigenbasis components revealed by image compression}

The rationale behind using principal component analysis for dimensionality reduction relies on the assumption that the amount of perceptually relevant information correlates with the variance of the parameters. This association between variance and perceptual importance, however, cannot be taken for granted, since even a simple rescaling of a parameter is enough to change its variance while the amount of perceptually relevant information remains invariant under invertible transformations. Thus, a rigorous approach to determining the perceptual importance of different parameters necessitates searching for an optimal quantization, as it is done in lossy image compression. 

The objective evaluation of lossy image compression algorithms is very difficult. In order to convince ourselves that the (quantized) eigenbasis components constitute a meaningful representation of naturally occurring textures, we set out to compare its coding efficiency with a popular compression algorithm such as JPEG 2000. For JPEG 2000 we can directly determine how many bits are required to encode an image with a certain mean squared reconstruction error. For our texture representation this is not straightforward, because the mean squared error between two randomly selected instantiations of the same texture can be very large. However, we can first build an encoder of the low-pass residual and then try to fill-in the high-frequency information based on the statistical constraints imposed by the texture parameters using the low-pass residual as an initialization. 
Thus, we have eliminated the randomness in the texture generation process and, instead, use the texture information to obtain a deterministic reconstruction of the pixel intensities.


Our previous finding that keeping the first 200 principal components is sufficient to obtain a perceptually faithful representation is equivalent to saying that the remaining parameters are assigned only one quantization bin each and thus require zero bits for their encoding. However, determining how many bins are required for each of the first 200 principal components in general, is a painstaking engineering task which ultimately can be solved only by a combination of heuristics with trial-and-error. 

Since most texture parameters are transformed covariances estimated from small image patches, the parameter values carry a fairly high level of uncertainty, suggesting that their numerical representation cannot require high precision. 
In fact, without considerable perceptual robustness with respect to small parameter deviations, texture analysis and subsequent resynthesis with the PS algorithm would not yield satisfying results. In the following, we sketch a quantization scheme yielding a performance comparable to JPEG 2000.

\begin{figure}[htbp]
\begin{center}
	\includegraphics[width=8.3cm]{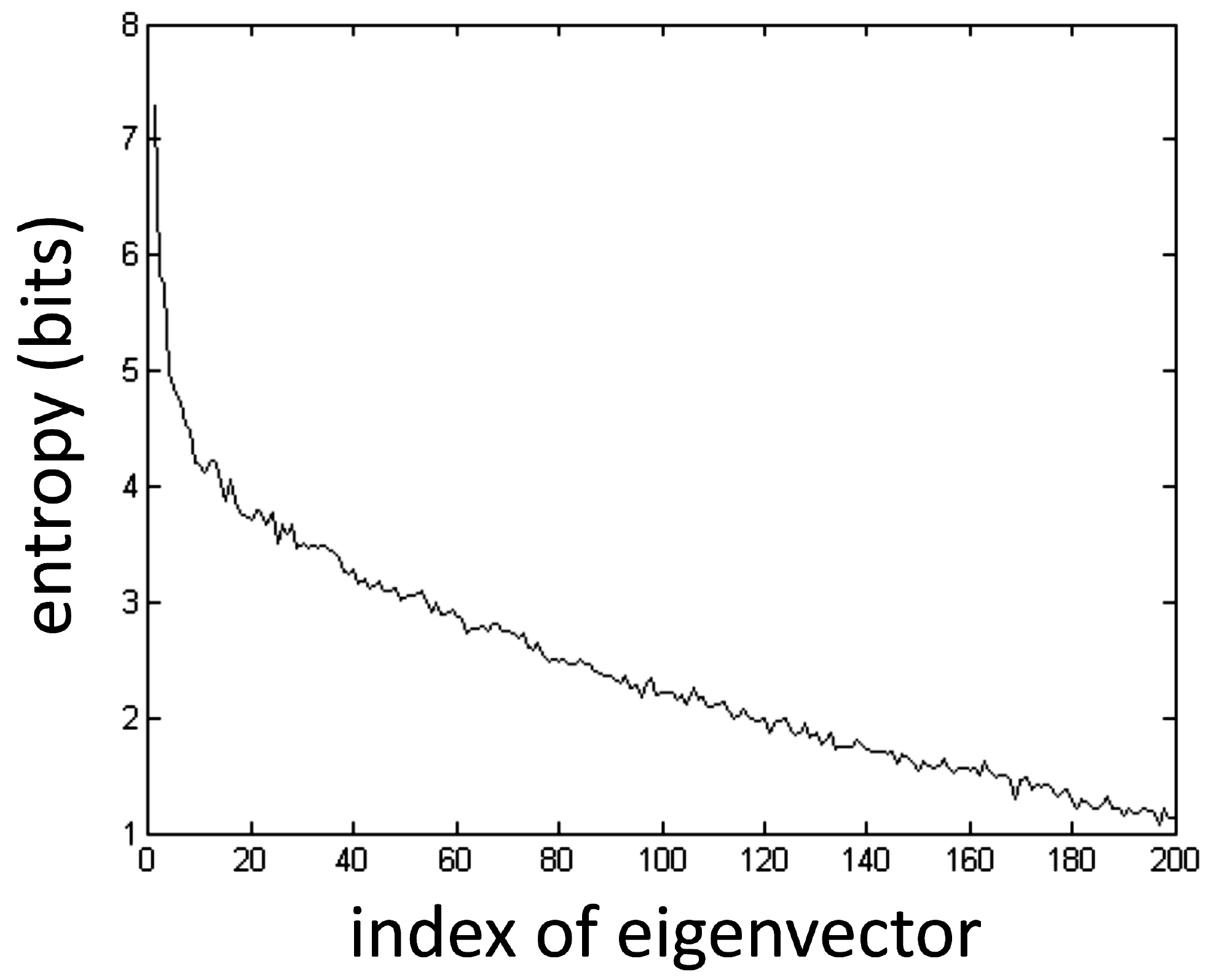}
\caption{
{\bf Bit allocation to the prinicipal components in texture parameter space.} Based on the empirical distribution of the coefficients of the texture parameter vector expressed in the eigenbasis, one can compute the entropy of each coefficient given a particular quantization. We base the choice of quantization of a coefficient on the corresponding eigenvalue, the underlying assumption being that the magnitude of an eigenvalue directly corresponds to its perceptual relevance, which is supported by empirical observation (see Fig.~\ref{dimensionalityReduction}).}
\label{bitAllocation}
\end{center}
\end{figure}

In order to quantize the coefficients of a texture parameter vector expressed in the natural eigenbasis, it is necessary to determine the marginal distributions of all coefficients across natural textures. The empirical marginal histograms can then be partitioned depending on the desired resolution.
Provided that the variance of a principal component correlates with its perceptual relevance, it can serve as a guide for the allocation of bits to each principal component. The previously established link between the number of perceptually required basis functions and the eigenvalues (cf.~Fig.~\ref{dimensionalityReduction}C\&D) suggests that this should be the case. Of course, the correspondence between perceptual relevance and eigenvalues had to be empirically verified  in more detail.
Therefore, we manually adjusted the quantization of the principal components by minimizing the perceived difference between compressed image and original. From the empirical distribution of the eigenvector coefficients across natural image patches we obtain the corresponding entropies, indicating the minimum number of bits required to encode the coefficients at the chosen level of quantization (Fig.~\ref{bitAllocation}). 



Compression of the low-pass filtered initialization image was achieved by applying a 2D discrete cosine transform (DCT), setting the coefficients of high-frequency components to zero, quantizing the remaining coefficients, and subsequent back-transforming. The frequency threshold was determined heuristically based on a perceptual judgement of image quality in terms of the preservation of long-range object contours. We found that the first ten eigenvectors should be allocated a much higher bit resolution than the remaining ones. 

Figure \ref{dimensionalityReduction}E  shows a quantized version of the original image in Figure \ref{dimensionalityReduction}A. The same quantization scheme was applied to the texture parameter vector of each patch, resulting in a representation requiring only 0.1376 bits/pixel (including the low-frequency initialization). 
For comparison, we show a JPEG2000 encoding that has similar compression rate (Fig.~\ref{dimensionalityReduction}F, encoded at 0.1389 bits/pixel). 
The two compression methods are somewhat complementary. While the JPEG2000 algorithm maintains contours at the expense of texture quality and introducing some local artifacts (spurious ``Gabor-patches''), our compression scheme renders the different textures quite well, though with some loss of long-range contour fidelity. 

Our comparison with JPEG 2000 shows that the first 200 components of the natural texture eigenbasis are perceptually highly informative features. It also provides the basis for quantitative studies of perception of crowing and visual search  \cite{Rosenholtz} and modeling of the underlying neural representation,  which is likely to be located in V2 \cite{FreemanEtAl_V2signature}. Beyond its implications for the study of biological vision, its competitive performance in low-rate image compression may also influence the design of new image compression algorithms.


\subsection*{Perceptual Evaluation of Image Models}


To asses the relation between neural response properties and the perceptual precision with which images are encoded it is necessary to know the perceptually relevant statistics of natural images. Since it is much easier to measure the statistics of pixel intensities than the perceptual representation of images, most neural image representation models only account for the statistics of pixel intensity patterns but are not informed by perceptual measurements. Thus, it is an important open question how well these models can explain the accuracy with which differences between images or image ensembles can be assessed perceptually. One possibility of course is to perform psychophysical experiments \cite{Gerhard2013a}.

However, since psychophysical evaluations are very time-consuming and costly, it is desirable to detemine statistical features that are known to capture perceptually important information, and thus can provide important hints about how much a model has captured perceptually relevant information.
Given the perceptual importance of the Portilla-Simoncelli texture parameters, one can now use them as a reference to measure the extent to which a statistical image model of interest is able to reproduce their statistics. For this purpose, a set of images sampled from the model under consideration is analyzed in the same manner as the natural image ensemble, in order to obtain the texture parameter mean and covariance matrix, i.e.~the Gaussian meta-statistics of the image model. 

\begin{figure}[h]
\begin{center}
 \includegraphics[width=12.35cm]{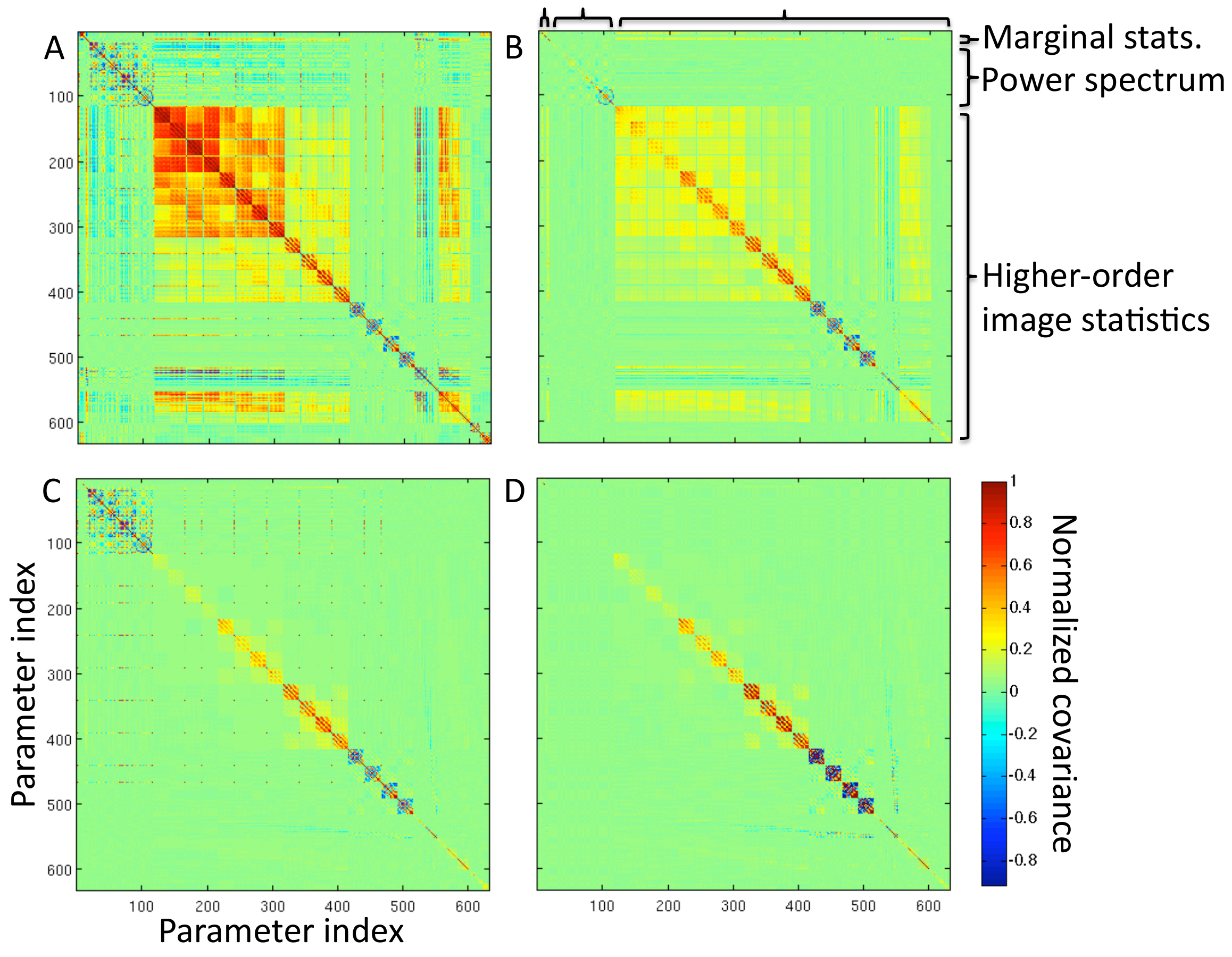}
\caption{
{\bf Parameter covariance matrices obtained from different image ensembles.} A: natural images. B: samples from the MCGSM image model (see text). C: phase-scrambled natural images. D: Gaussian noise with averaged power spectrum of natural images. 
each covariance matrix was normalized with respect to the diagonal elements of the {\em natural} covariance matrix serving as a common reference.
}
\label{covarianceMatrices}
\end{center}
\end{figure}

{\bf In comparing visualizations of the covariance matrices, one can obtain an intuitive overview of the model's quality ``by inspection.'' 
Figure \ref{covarianceMatrices}A shows the correlation matrix obtained from natural image patches, i.e. the covariance matrix normalized with respect to its diagonal elements (variances). Panels B through D were obtained from three image models in decreasing order of complexity. Again, each covariance matrix was normalized with respect to the diagonal elements of the {\em natural} covariance matrix, which thus serves as a common reference for the scaling of variances and covariances for all image models. }
The mixture of conditional Gaussian scale mixtures (MCGSM) approach \cite{MCGSM} is a sophisticated pixel-based model able to capture a substantial amount of the cross-orientation and cross-scale dependencies (off-diagonal block structure) among those parameters encoding higher-order image correlations (Fig.~\ref{covarianceMatrices}B); though, in comparison with A, cross-scale parameter correlations tend to be lower than intra-scale correlations. Figure \ref{covarianceMatrices}C was obtained by analyzing phase scrambled natural image patches. As expected, this image model fully captures the power spectrum, but nothing beyond correlations of magnitude autocovariances within the same respective orientation and scale (diagonal blocks).

Note that phase scrambling within the different images only destroys the higher-order correlations within each individual image. Thus, the statistics obtained for the ensemble of many different phase scrambled images cannot be understood as a single Gaussian distribution but corresponds to a mixture of Gaussians where one first samples a power-spectrum (or equivalently an auto-correlation function) and then generates a Gaussian random field. For comparison, we also study the case where we describe natural images by a single Gaussian process rather than a mixture of Gaussian processes (Fig.~\ref{covarianceMatrices}D). To this end, we always have to use the same power spectrum instead of using different power spectra for each image. Therefore, in a simple Gaussian process model the sampled textures share only the average power spectrum with natural images. Hence, the Fourier coefficients have the same mean value as observed in natural image patches, but the coefficient variances and covariances are not captured. Rather, the coefficient values are independent Gaussian random variables with equal variance. 
{ \bf Since in this model there is no structure beyond the power spectrum the covariance matrix should be zero.
 Any residual dependencies observed in practice (Fig.~\ref{covarianceMatrices}D) we attribute to measurement errors due to finite patch size of only $64\times64$ pixels. When applied to such small image patches, the phase scrambling operation is imperfect, leading to inaccuracies in the estimation of baseline HOC-parameters. 
 While we reduce the error by averaging baseline HOC-parameter values obtained from repeated phase scrambling with different phase randomizations, some erroneous correlations remain.
 } 

Although not all higher-order correlations in natural textures can be captured by the correlations of filter coefficient magnitudes, these limitations in descriptive power do not undermine the validity of choosing the Portilla-Simoncelli texture parameters as a basis for comparing image models. 
The comparison is made with respect to this particular reference set of image statistics, chosen based on their empirically determined perceptual relevance. Thus, we have established a form of benchmark test for image models.

\subsubsection*{Image model comparison in the natural eigenbasis}

In addition to the visual comparison of hyperparameters described above, the difference in the meta-statistics of a given image model and natural images can be expressed in terms of the Kullback-Leibler divergence (KLD) between the two underlying probability densities.
Let  $\mathcal{N}_0 ( \mu_0, \Sigma_0)$ be the multivariate Gaussian probability density describing the second-order meta-statistics of natural images (reference density); $\mathcal{N}_1 (\mu_1, \Sigma_1)$, the corresponding density of a given image model; and $n$, the number of dimensions (texture parameters). Then their Kullback-Leibler divergence,
\begin{equation}
D_{KL}( \mathcal{N}_0 || \,\mathcal{N}_1) = \frac{1}{2} \left[  \mbox{tr}(\Sigma_1^{-1} \Sigma_0)  +   
(\mu_1 - \mu_0)^T  \Sigma_1^{-1}  (\mu_1 - \mu_0) - \ln \left(  \frac{\det \Sigma_0}{ \det \Sigma_1} \right) - n
   \right] ,
   \label{KLD}
\end{equation}

\noindent
measures the deviation of model meta-statistics from the natural reference. Numerically, this quantity remains
computable even for a large number of very small eigenvalues, though the accuracy can then be affected by error accumulation. In the next section, we demonstrate how equation \ref{KLD} can provide a quantitative means of 
comparing image models in the natural eigenbasis.

\subsubsection*{Subspace principal component analysis}

By virtue of the transformations applied in our texture analysis scheme (Fig.~\ref{analysisScheme}), parameters representing the power spectrum are largely independent of those encoding higher-order image correlations, as evidenced in the block structure of the parameter covariance matrix (Fig.\ref{covarianceMatrices}A).
Therefore, it is advantageous to diagonalize the corresponding subspaces separately (subspace PCA).
For each of the two subspaces, we perform the comparison between image model and natural reference 
by projecting both model and natural covariance matrices onto a variable subset of the natural eigenbasis and by subsequently evaluating (\ref{KLD}). The KLD is then plotted as a function of the number of eigenvectors taken into account. Thus, we can separately evaluate image model performance with respect to second-order and higher-order correlations, respectively.

\begin{figure}[ht]
\begin{center}
   \includegraphics[width=12.35cm]{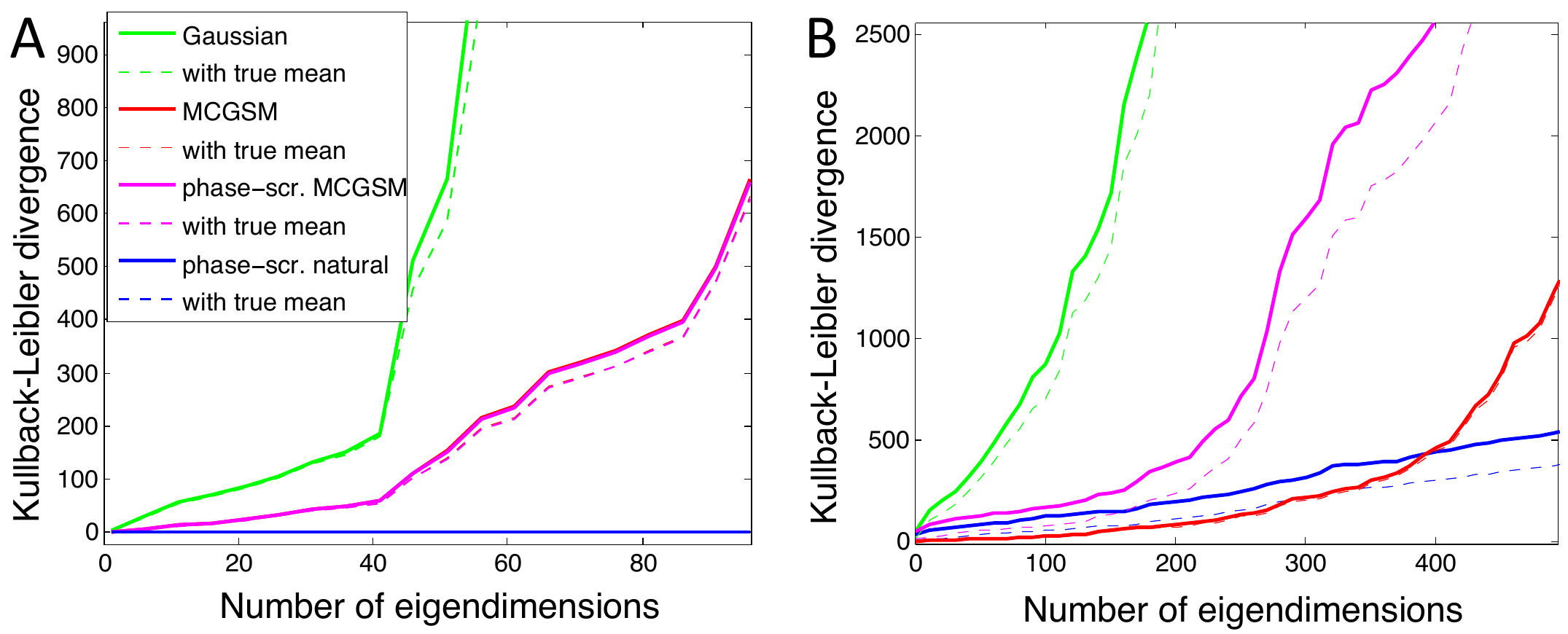}
\caption{
{\bf Image model comparison via the Kullback-Leibler divergence (KLD) of model and natural meta-statistics.} 
The parameter space has been partitioned into two subspaces, one comprising the parameters encoding the power spectrum (A) and the other, the parameters encoding higher-order correlations (B). 
For each subspace, the KLD is plotted as a function of the number of eigenvectors taken into account. 
Three models are evaluated: phase-scrambled natural images, which capture only the second-order statistics (solid blue curve); the mixture of conditional Gaussian scale mixtures (MCGSM) model (solid red curve); and Gaussian noise with an average power spectrum matched to natural images (solid green curve). The dotted curves are KLD plots for the same models but with the model mean parameter vectors substituted by the true (i.e.~the natural) mean parameter vector (setting $\mu_1 = \mu_0$ in Eqn.~\ref{KLD}), which therefore is a comparison based solely on the parameter covariance matrices. 
A perfect image model yields a straight line at zero, since the second-order meta-statistics of its samples must be identical to that of natural image patches. This is the case for phase-scrambled natural images in panel A (coinciding blue curves), since the second-order statistics is--by construction--completely preserved.
 In addition, we analyzed phase-scrambled MCGSM samples (magenta curves). Since this model does not fully capture the power spectrum of natural image patches, removing the higher-order correlations makes the MCGSM-samples inferior to phase-scrambled natural patches. By definition, the curves for MCGSM samples (red) and phase-scrambled MCGSM samples (magenta) in panel A coincide.}
\label{KLDplot}
\end{center}
\end{figure}

Figure \ref{KLDplot} shows this KLD comparison for the three previously described image models: MCGSM (red), phase-scrambled natural patches (blue), and Gaussian noise with mean natural power spectrum (green). The dotted curves are plots of the KLD computed based on the true mean (i.e.~by setting $\mu_1 = \mu_0$, the mean parameter vector over natural images). Through this substitution one can distinguish the contribution of mean vector and covariance matrix in the performance comparison.

As expected, the MCGSM model is clearly superior to the pink noise images (phase-scrambled natural images), and the Gaussian noise model performs worst.
The reduction in KLD induced by substituting the true mean parameter vector of natural images (i.e.~the difference between solid and dashed curves) is minute for the MCGSM model, indicating that its mean parameter vector is closer to that of natural images. For the same reason, when the pink noise model (blue) is equipped with the natural mean vector, the distance to the natural meta-statistics is reduced to about the same level. Hence, the dotted red and blue curves in Figure \ref{KLDplot}B are highly similar, except for the latter part of the graph, where the KLD 
is increasingly dominated by accumulating inaccuracies of many and extremely small eigenvalues (cf.~Fig.\ref{cumVariance}). Since the eigenvectors corresponding to very small eigenvalues tend to be perceptually irrelevant, the KLD comparison should be limited to the eigen-subspace of the significant principal components.


\section*{Discussion}

We have built a generator of natural texture surrogates based on the empirical distribution of the Portilla-Simoncelli texture parameters across natural image patches. The main contribution of this work is that we established a new representation of the texture parameters which precludes  inadmissible parameter combinations. This new representation sets the stage for many important applications. For example, it is now possible to randomly draw textures, as required for system identification experiments. The importance of Portilla Simoncelli textures for probing the visual system has been recently demonstrated when studying differences in the responses of V1 and V2 neurons \cite{FreemanEtAl_V2signature}. In addition, the presence of controllable higher-order correlations in our stimuli could also aid the characterization of receptive field non-linearities via the estimation of Wiener kernels \cite{YuEtAl_nonlinearReverseCorrelation_2003}.

Our approach is fundamentally different from that of Rust and Di Carlo \cite{rustDiCarlo_PSscrambling_2010}, who have previously applied the PS algorithm to obtain scrambled versions of natural image stimuli. Since in that study, each natural stimulus contained multiple textures, the re-synthesized surrogate was effectively a reconstruction of the \emph{average} over the textures. Of course, the authors specifically intended to remove certain features of natural stimuli (scrambling), rather than to model natural image statistics.
In contrast, our stimuli are based on the analysis of multiple image patches that are small enough and especially selected to contain mostly one type of texture in each instance. Therefore, the stimuli generated with our method retain more of the correlations characteristic of natural images.

There are two main factors limiting the statistical complexity of our surrogate textures: the meta-statistical model and the descriptive power of the texture parameters. In comparison with textures sampled from our model, resynthesized natural textures are more similar to actual natural textures. Hence, our model of the joint parameter distribution does not fully exhaust the potential of the Portilla-Simoncelli algorithm. The question then arises as to how the meta-model fails to fully describe the parameter dependencies. Though it is difficult to exhaustively investigate all possible parameter combinations, spot checks in the form of comparing scatter plots of selected parameter pairs produced the empirical finding that the multivariate Gaussian model appears to capture the \emph{pairwise} dependencies of the transformed parameters quite well (results not shown). Consequently, we attribute the greater visual complexity of resynthesized natural patches not so much to shortcomings in the modeling of pairwise dependencies but rather to the existence of significant \emph{higher-order} parameter correlations. 

A more complex meta-model, such as a mixture of multivariate Gaussians, could provide a more accurate description of the parameter dependencies and yield even more naturalistic textures. Our methodology for achieving intrinsic parameter consistency would ensure that samples from any such meta-model would be converted into valid texture parameters. On the other hand, the fact that natural textures cannot always be faithfully reproduced with the PS-algorithm demonstrates the incompleteness of the feature set employed by the algorithm. Further image statistics could be incorporated into the texture algorithm, without having to change the overall concept of our approach. For instance, in their original paper, Portilla and Simoncelli \cite{PSalg} mention the absence of end-stopping in their algorithm. However, while there is good reason for wishing to include end-stopping, it is not clear how to implement this type of feature, especially in the synthesis procedure.

The principal components of the texture parameters obtained from natural image patches form what we termed the \emph{natural eigenbasis}. We quantitatively  investigated the perceptual importance of the eigenbasis components. First we showed that the number of texture parameters can be reduced by more than a factor of three.
In conjunction with a tiling scheme inspired by the work of Rosenholtz and colleagues \cite{Balas_lowpass, Rosenholtz_textureTilingModel}, this led to a method of lossy image compression in the form of a patch-by-patch decomposition and encoding of an image. The local texture in each patch is encoded in a very compact manner via a representation in the dimensionality-reduced natural eigenbasis with quantized coefficients, requiring less than 0.14 bits/pixel. A great remaining challenge is posed by the sharp long-range object contours in natural scenes, such as blades of grass, which are not well reproduced, even though they clearly form a texture. The steerable filters employed in the PS algorithm are probably not optimally suited for this type of image feature. 
 
Finally, we have demonstrated how evaluations of the meta-statistics on the texture parameters can be useful for the comparison of natural image models. More specifically, we have shown how one can quantify the degree to which this model captures the pairwise dependencies of the Portilla-Simoncelli texture statistics in natural images, in terms of the Kullback-Leibler divergence. The comparison can also visualized by the parameter covariance matrices obtained from natural and model images. Though this is a comparison only with respect to the pairwise dependencies of a specific set of texture features, it nevertheless constitutes a significant benchmark test that allows efficient comparison of models with respect to perceptually important features.

In summary, the development of the new representation for the Portilla Simoncelli texture parameters opens many new possibilities for studying neural and perceptual properties of the visual system. We have only started to explore some of these possibilities here but envision further improvements and also the inception of new applications in the future.






\section*{Acknowledgments}


\bibliography{texturePaper}

\section*{Figure Legends}

\begin{figure}[!ht]
\begin{center}
\includegraphics[width=8.3cm]{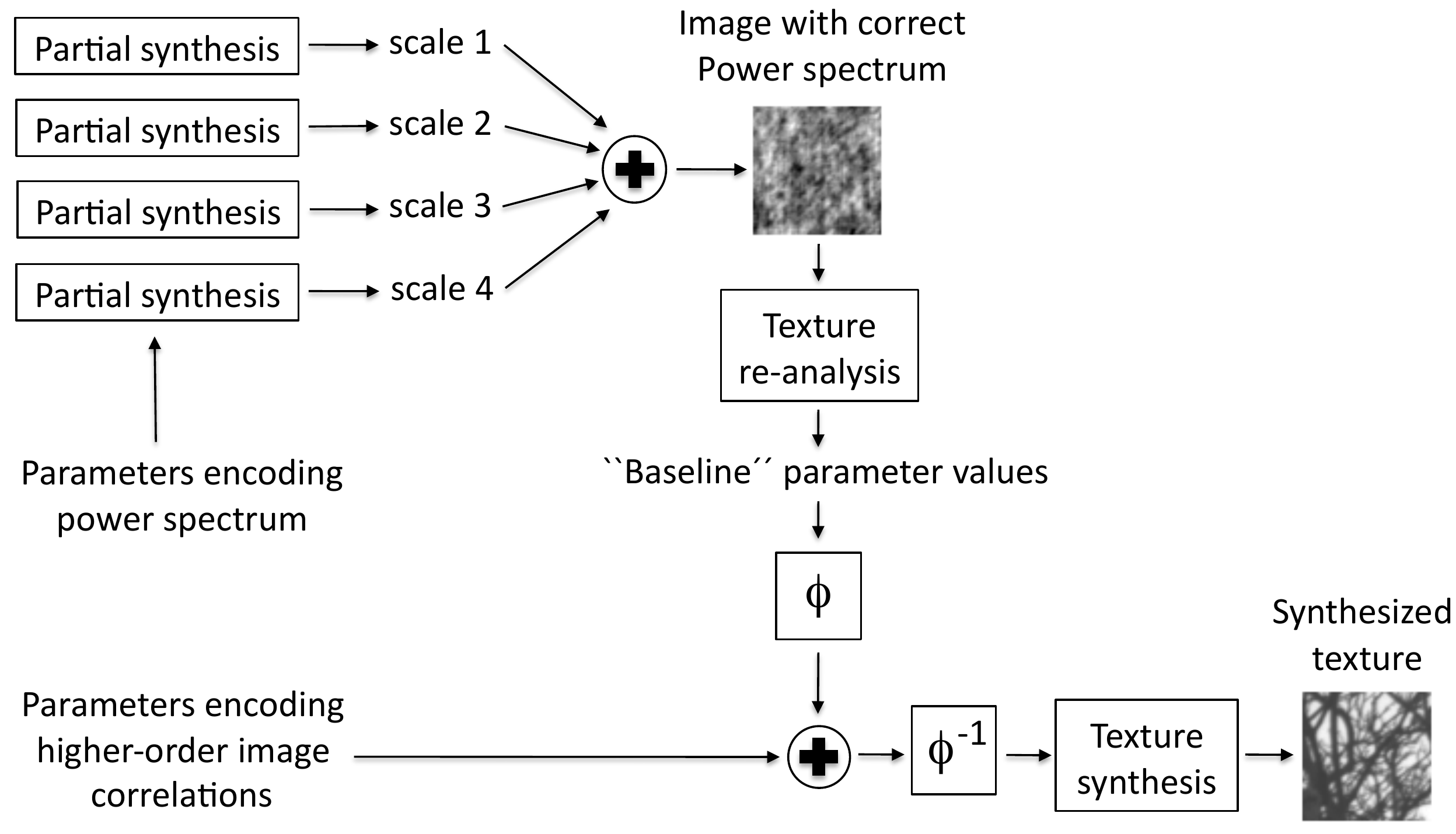}
\caption{
{\bf Schematic diagram of our texture synthesis procedure.} Since the Gaussian meta-model captures only pairwise parameter relationships, the ``raw" parameter values drawn from the model lack simultaneous consistency across space and scale. Therefore, the second-order correlation structure (power spectrum) is established by partial texture reconstruction at each scale. Reanalyzing the sum of the resultant images from all scales yields a new set of scale-consistent parameters, encoding the second-order image statistics as well as the baseline values of the remaining higher-order parameters. The transformation $\phi$ subsumes all symmetrizing parameter transformations, the normalization of covariances to correlations, and the Cholesky-decomposition of those parameter groups that are filter response covariance matrices. Finally, adding the baseline values to the higher-order parameters drawn from the meta-model yields a consistent full parameter vector suitable for texture synthesis.}
\label{synthesisScheme}
\end{center}
\end{figure}

\end{document}